\documentclass[letterpaper]{article} % DO NOT CHANGE THIS
\usepackage{aaai2026}  % DO NOT CHANGE THIS
\usepackage{times}  % DO NOT CHANGE THIS
\usepackage{helvet}  % DO NOT CHANGE THIS
\usepackage{courier}  % DO NOT CHANGE THIS
\usepackage[hyphens]{url}  % DO NOT CHANGE THIS
\usepackage{graphicx} % DO NOT CHANGE THIS
\usepackage{natbib}  % DO NOT CHANGE THIS AND DO NOT ADD ANY OPTIONS TO IT
\usepackage{caption} % DO NOT CHANGE THIS AND DO NOT ADD ANY OPTIONS TO IT
\usepackage{algorithm}
\usepackage{algorithmic}
\usepackage{multirow}
\usepackage{amsmath}
\usepackage{newfloat}
\usepackage{listings}
\DeclareCaptionStyle{ruled}{labelfont=normalfont,labelsep=colon,strut=off} % DO NOT CHANGE THIS
\floatstyle{ruled}
\newfloat{listing}{tb}{lst}{}
\floatname{listing}{Listing}
\usepackage{amssymb}

\title{V2VLoc: Robust GNSS-Free Collaborative Perception via LiDAR Localization}
\author{
    Wenkai Lin\textsuperscript{\rm 1,\rm 2}\equalcontrib,
    Qiming Xia\textsuperscript{\rm 1,\rm 2}\equalcontrib,
    Wen Li\textsuperscript{\rm 1,\rm 2},
    Xun Huang\textsuperscript{\rm 1,\rm 2},
    Chenglu Wen\textsuperscript{\rm 1,\rm 2}\thanks{Corresponding author.} \\
}
\affiliations{
    \textsuperscript{\rm 1}Fujian Key Laboratory of Sensing and Computing for Smart Cities, Xiamen University, China\\
    \textsuperscript{\rm 2}Key Laboratory of Multimedia Trusted Perception and Efficient Computing\\
    
}
\usepackage{bibentry}
\begin{document}

\maketitle

\begin{abstract}
Multi-agents rely on accurate poses to share and align observations, enabling a collaborative perception of the environment. However, traditional GNSS-based localization often fails in GNSS-denied environments, making consistent feature alignment difficult in collaboration. To tackle this challenge, we propose a robust GNSS-free collaborative perception framework based on LiDAR localization. Specifically, we propose a lightweight Pose Generator with Confidence (PGC) to estimate compact pose and confidence representations. To alleviate the effects of localization errors, we further develop the Pose-Aware Spatio-Temporal Alignment Transformer (PASTAT), which performs confidence-aware spatial alignment while capturing essential temporal context. Additionally, we present a new simulation dataset, V2VLoc, which can be adapted for both LiDAR localization and collaborative detection tasks. V2VLoc comprises three subsets: Town1Loc, Town4Loc, and V2VDet. Town1Loc and Town4Loc offer multi-traversal sequences for training in localization tasks, whereas V2VDet is specifically intended for the collaborative detection task. Extensive experiments conducted on the V2VLoc dataset demonstrate that our approach achieves state-of-the-art performance under GNSS-denied conditions. We further conduct extended experiments on the real-world V2V4Real dataset to validate the effectiveness and generalizability of PASTAT. 
% Our dataset and code will be made publicly available.
\end{abstract}
% Uncomment the following to link to your code, datasets, an extended version or similar.
% You must keep this block between (not within) the abstract and the main body of the paper.
\begin{links}
    \link{Code}{https://github.com/wklin214-glitch/V2VLoc}
    \link{Datasets}{https://huggingface.co/datasets/linwk/V2VLoc}
\end{links}

\section{Introduction}

Collaborative perception is a paradigm in which multi-agents share information and cooperate on perception tasks to achieve improved accuracy, wider coverage. Recently, collaborative perception has experienced rapid development~\cite{where2comm, codefilling, traf-align} and has demonstrated promising performance.

The primary capability of collaborative perception lies in its ability to integrate observations from various viewpoints into a unified coordinate frame. This integration process is generally reliant on precise pose estimation. Recent studies \cite{v2x-vit, CoAlign,ermvp} have attempted to improve robustness by mitigating the effects of localization errors. However, traditional GNSS-based localization methods, such as GPS combined with RTK receivers, are inherently device-dependent and are prone to degraded or lost signals in GNSS-denied environments (e.g., tunnels, spoofing, jamming, or satellite occlusion). This dependency presents significant challenges for effective collaborative perception.

\begin{figure}[t]
\centering
\includegraphics[width=1.0\columnwidth]{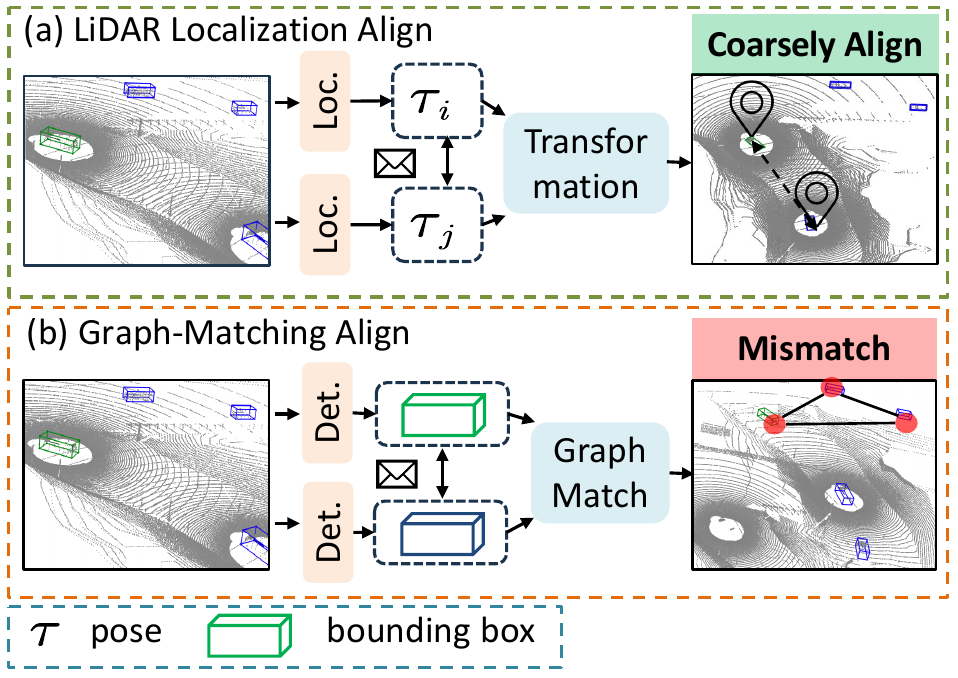} % Reduce the figure size so that it is slightly narrower than the column. Don't use precise values for figure width. This setup will avoid overfull boxes.
\caption{An illustration of different alignment methods. (a) shows that LiDAR localization achieves coarse alignment successfully. (b) shows that the Graph-Matching method fails without consensus objects. }
\label{fig1}
\end{figure}

To tackle this issue, FreeAlign~\cite{freealign} proposed an alignment method that utilizes the intrinsic geometric patterns present in the sensor data for inter-agent alignment and relative pose estimation. However, as illustrated in Fig. \ref{fig1}(b), this graph-matching-based method relies on bounding box sharing, pairwise greedy graph matching, and assumes the presence of a certain number of co-viewed objects, which makes it less effective in scenarios with sparse or minimally overlapping observations, ultimately leading to unstable pose alignment.

Recent advances have been made in LiDAR-based visual localization models~\cite{choy2019fully, xia2021soe, zhang20233d, sgloc,lisa,diffloc,nidaloc,lightloc},  which estimate global coordinates using LiDAR data. Compared to map-based methods with heavy storage and communication costs, regression-based LiDAR localization offers a lightweight and scalable solution.
% We observe that regression-based LiDAR localization demonstrates favorable performance, as shown in Tab. \ref{intro1}. 
Inspired by it, we propose a novel GNSS-free collaborative perception framework that uses a lightweight regression-based LiDAR localization module to transmit pose and confidence representations, thereby reducing bandwidth, latency, and alignment failures (see Tab. \ref{intro1}).

\begin{figure}[t]
\centering
\includegraphics[width=1.0\columnwidth]{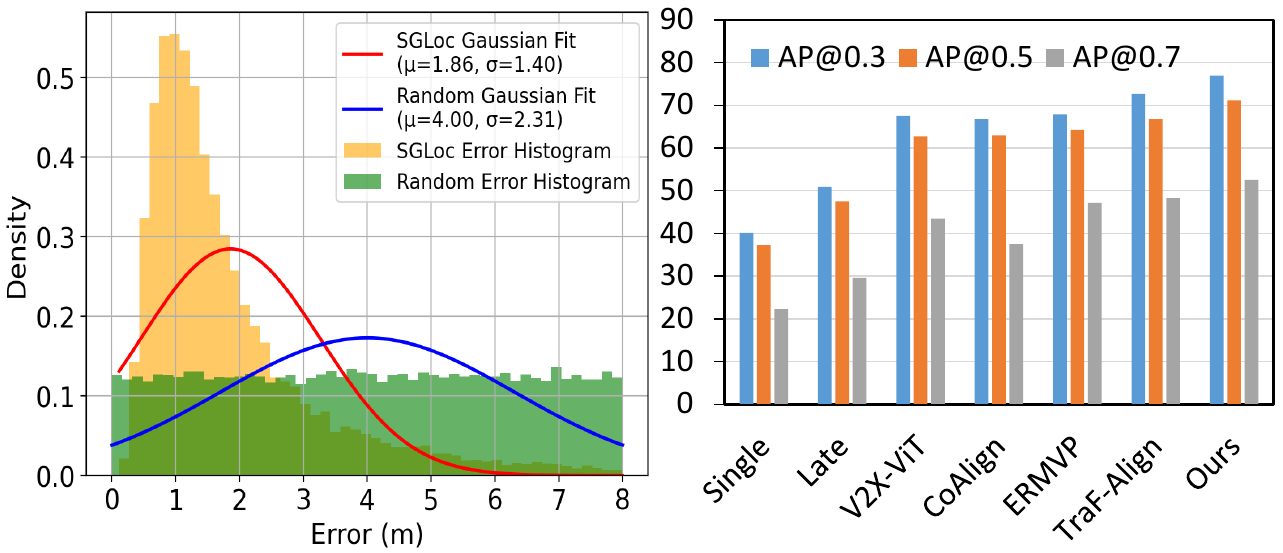} % Reduce the figure size so that it is slightly narrower than the column. Don't use precise values for figure width. This setup will avoid overfull boxes.
\caption{
The left figure shows the error distribution of SGLoc \cite{sgloc} and random noise at a noise level of 8.0 (m); the right shows model performance on V2VDet under LiDAR localization-induced pose errors.
% Error distribution histogram and Gaussian curve of SCR and random noise.
}
\label{fig2}
\end{figure}

% Our experiments show that PGC achieves an inference time as low as 0.0068 seconds, enabling efficient and scalable multi-agent collaboration.
However, several challenges remain. One major challenge is the lack of datasets that support both collaborative perception and regression-based LiDAR localization, which requires separate traversals of the same environment for training and testing. Another challenge lies in localization errors caused by imperfect pose estimation. Unlike previous methods \cite{where2comm, CoAlign, ermvp} that simulate localization noise using synthetic Gaussian perturbations to verify the robustness of their methods, our approach captures localization errors that naturally emerge from intricate and structured environmental conditions. As shown in Tab. \ref{table3} and the left figure of Fig. \ref{fig2}, detection performance differs between random noise and LiDAR localization errors, as random noise distributions fail to accurately model the structured and environment-driven characteristics of pose estimation errors.

Therefore, we propose a novel simulated dataset named V2VLoc. The V2VLoc dataset is composed of three subsets: Town1Loc, Town4Loc, and V2VDet. 
As illustrated in Fig.~\ref{method}, we also develop a two-stage collaborative detection framework: 
% (i) the single-agent Pose Generator with Confidence (PGC) model, which is trained on the Town1Loc and Town4Loc subsets; and (ii) per-frame pose and confidence estimation, followed by feature alignment and multi-agent collaborative detection on the V2VDet subset. 
(i) we train a single-agent Pose Generator with Confidence (PGC) model on the Town1Loc and Town4Loc subsets; and (ii) we perform per-frame pose and confidence estimation, followed by feature alignment and multi-agent collaborative detection on the V2VDet subset.
% The details of V2VLoc  will be discussed in the Experiments section.

\begin{table*}[t]
\centering
% \resizebox{1.0\textwidth}{!}{
\renewcommand{\arraystretch}{1.1}
\begin{tabular}{l|cccccccc}
\hline
Dataset    & Year & R/S & Sensor  & V2X    & Image (360°) & Agent Number & 3D Boxes  & Multi-traversal \\ \hline
% KITTI     & 2012 & R   & L\&C    & ×      & ×           & 1            & \checkmark                              & × 
%               \\
% NCLT       & 2016 & R   & L\&C    & ×      & ×           & 1            & ×                               & \checkmark
%               \\
% Oxford    & 2017 & R   & L\&C\&R & ×      & ×           & 1            & ×                             & \checkmark 
%                \\ \hline
OPV2V   & 2022 & S   & L\&C    & V2V    & \checkmark           & 7          & \checkmark                             & ×               \\
V2X-Sim   & 2022 & S   & L\&C    & V2V\&I & \checkmark           & 10           & \checkmark                             & ×               \\
V2XSet  & 2022 & S   & L\&C    & V2V\&I & \checkmark           & 5          & \checkmark                              & ×               \\
DAIR-V2X  & 2022 & R   & L\&C    & V2I    & ×           & 2            & \checkmark                          & ×               \\
V2V4Real   & 2023 & R   & L\&C    & V2V    & ×           & 2            & \checkmark                            & ×               \\
V2X-Seq & 2023 & R   & L\&C    & V2I    & ×           & 2            & \checkmark                            & ×               \\
V2XReal    & 2024 & R   & L\&C    & V2V\&I & \checkmark           & 4            & \checkmark                              & ×               \\
V2X-Radar  & 2024 & R   & L\&C\&R & V\&I   & ×           & 2            & \checkmark                            & ×               \\
Open Mars  & 2024 & R   & L\&C       & V2V    & ×           & 3          & ×                            & \checkmark               \\ 
V2X-R    & 2025 & S   & L\&C\&R & V2V\&I    & \checkmark           & 5          & \checkmark                             & ×               \\
\hline
% V2VLoc     & - &- & - & -   &      -  &    -     &   -        &     -          \\
V2VLoc (Ours)     & 2025 &S & L\&C\&R & V2V   &  \checkmark           & 4            & \checkmark                              & \checkmark      \\
  - Town1Loc & 2025 & S   & L\&C\&R &  ×   & \checkmark           & 1            & \checkmark                              & \checkmark               \\
  - Town4Loc & 2025 & S   & L\&C\&R & ×    & \checkmark           & 1            & \checkmark                            & \checkmark               \\
  - V2VDet   & 2025 & S   & L\&C\&R & V2V    & \checkmark           & 4          & \checkmark                              & ×               \\ \hline
\end{tabular}
% }
\caption{Comparison of different datasets including OPV2V \cite{opv2v}, V2X-sim \cite{v2x-sim}, V2XSet \cite{v2x-vit}, DAIR-V2X \cite{dair-v2x}, V2V4Real \cite{v2v4real}, V2X-Seq \cite{v2x-seq}, V2XReal \cite{V2x-real}, V2X-Radar \cite{v2x-radar2024v2x}, Open Mars \cite{openmars2024multiagent}, V2X-R \cite{v2x-R} and V2VLoc. R/S: Real-world/Simulated, C: Camera, L: Lidar, R: Radar, V2V: vehicle-to-vehicle, V2I: Vehicle-to-infrastructure. }
% Oxford \cite{oxford20171}, KITTI \cite{kitti2012we}, NCLT \cite{nclt2016university},
\label{t1}
\end{table*}

 We also propose a novel module, Pose-Aware Spatio-Temporal Alignment Transformer (PASTAT), to alleviate the impact of localization errors. We first perform coarse alignment of features derived from various viewpoints, and incorporate the pose confidence as a learnable Confidence Embedding (CE). Subsequently, the coarsely matched features are further refined by the Feature Spatial Alignment (FSA) module for more accurate alignment.  A Vision Transformer with Temporal Encoding (TE) is then applied to capture spatio-temporal context across multiple frames. Finally, the detection head processes the temporally encoded features to generate the final predictions.

To evaluate our method, we conduct extensive experiments on both V2VLoc and the real-world dataset V2V4Real \cite{v2v4real}. The results demonstrate that our method achieves the best performance in collaborative perception under GNSS-denied conditions, as shown in the right figure of Fig. \ref{fig2}, and also exhibits good performance in real-world scenarios. Our main contributions are summarized as follows:
\begin{itemize}

\item We are the first to apply LiDAR localization to solve the feature alignment problem in collaborative detection without GNSS signals. We establish a new perception pipeline that supports various fusion strategies and advances research on GNSS-free collaborative perception.
\item We propose V2VLoc, a novel simulation dataset that supports both regression-based LiDAR localization and multi-agent collaborative object detection tasks, which lays a solid data foundation for collaborative perception in GNSS-denied scenarios.

\item We design two key modules: the Pose Generator with Confidence (PGC) module and the Pose-Aware Spatio-Temporal Alignment Transformer (PASTAT) module to alleviate the impact of localization errors. Our method achieves superior performance across both the V2VLoc dataset and real-world benchmarks.

\end{itemize}

\section{Related Work}

\subsubsection{Multi-Agent Collaborative Perception.}
Multi-agent collaborative perception ~\cite{v2vnet, DiscoGraph, opv2v, cobevt, CoAlign,  CoCa3D, codefilling, ermvp, traf-align,dota} has attracted increasing attention due to its potential to enhance the performance of individual agents significantly. By sharing sensory information among agents and transforming their observations into a unified coordinate system via ego poses, the field of view is broadened, effectively alleviating issues such as occlusion. 
% Existing work has proposed various solutions to address the challenges in collaborative perception systems, including limited communication bandwidth\cite{codefilling}, pose errors\cite{CoAlign, traf-align, DiscoGraph}, and transmission delays\cite{v2x-vit}, to improve system robustness. 
However, most existing methods rely on external localization devices (e.g., GNSS or RTK) to obtain ego poses for cooperation, leading to degraded performance in GNSS-denied environments. FreeAlign \cite{freealign} addresses this issue by estimating relative poses through graph matching. However, it heavily depends on the availability of multiple co-viewed objects.

% \subsubsection{Scene Coordinate Regression.}
% Scene Coordinate Regression (SCR) has been widely adopted for visual relocalization and has recently been extended to LiDAR-based settings for ego pose estimation. These methods predict dense 3D correspondences between local observations and a global scene, enabling robust pose estimation via RANSAC. For example, SGLoc \cite{sgloc} was the first to propose such a decoupled regression framework for reliable localization. LiSA \cite{lisa} further introduced semantic information to improve localization accuracy, while LightLoc \cite{lightloc} significantly reduced the number of model parameters, effectively addressing the issue of slow training. In the context of collaborative perception, SCR offers a promising solution for aligning agents without GNSS signals.

\subsubsection{LiDAR Localization.}
Map-based LiDAR localization \cite{choy2019fully, xia2021soe, zhang20233d} estimates poses by matching query points with a pre-built 3D map; however, it often incurs high storage and communication costs. To address the limitations of map-based localization, regression-based methods such as Absolute Pose Regression (APR) \cite{wang2021pointloc, wang2023hypliloc, diffloc} and Scene Coordinate Regression (SCR) \cite{sgloc, lisa, lightloc,peloc} have emerged. APR directly regresses global poses from scene features, while SCR estimates poses by predicting LiDAR-to-world correspondences and applying RANSAC \cite{ransac}. By eliminating the reliance on explicit maps, these methods provide a lightweight and flexible alternative, making them suitable for GNSS-denied scenarios in collaborative perception systems.

\subsubsection{Collaborative Perception Dataset.}
Multi-agent collaborative perception has received increasing attention in recent years, with several studies focusing on the exploration and collection of more diverse datasets to support various tasks. For example, V2V4Real \cite{v2v4real} is the first large-scale V2V dataset captured in real-world driving scenarios, while RCooper \cite{rcooper} introduces the pioneering large-scale dataset for roadside collaborative perception. V2X-R \cite{v2x-R} is the first simulated V2X dataset that integrates LiDAR, camera, and 4D radar modalities. However, existing collaborative perception datasets are exclusively collected from single-pass trajectories, rendering them unsuitable for regression-based LiDAR localization tasks. This limitation presents a significant challenge for studying model robustness under GNSS-denied conditions.
% For example, SGLoc \cite{sgloc} was the first to propose such a decoupled regression framework for reliable localization. LightLoc \cite{lightloc} significantly reduced the number of model parameters, effectively addressing the issue of slow training. In the context of collaborative perception, regression-based LiDAR localization offers a promising solution for aligning agents without GNSS signals.

% Scene Coordinate Regression (SCR) has been widely adopted for visual relocalization and has recently been extended to LiDAR-based settings for ego pose estimation. These methods predict dense 3D correspondences between local observations and a global scene, enabling robust pose estimation via RANSAC. For example, SGLoc \cite{sgloc} was the first to propose such a decoupled regression framework for reliable localization. LiSA \cite{lisa} further introduced semantic information to improve localization accuracy, while LightLoc \cite{lightloc} significantly reduced the number of model parameters, effectively addressing the issue of slow training. In the context of collaborative perception, SCR offers a promising solution for aligning agents without GNSS signals.

\section{V2VLoc Dataset}
\subsection{Overview of V2VLoc}
Since the training and testing of regression-based LiDAR localization methods require different traversals of the same environment, this condition is not satisfied in the existing collaborative detection dataset. Therefore, we propose V2VLoc, the first dataset that supports both regression-based LiDAR localization and detection tasks. As shown in Tab. \ref{t1}, V2VLoc is a comprehensive dataset consisting of three subsets: Town1Loc, Town4Loc, and V2VDet. All subsets include multimodal sensors, such as LiDAR, camera, and 4D radar. The dataset offers multi-traversal scans and 3D bounding box annotations for all frames, supporting both localization and detection tasks.
 % designed to support research in collaborative localization and perception under GNSS-denied or degraded conditions.

\subsection{Data Collection}
\begin{figure}[t]
\centering
\includegraphics[width=1.0\columnwidth]{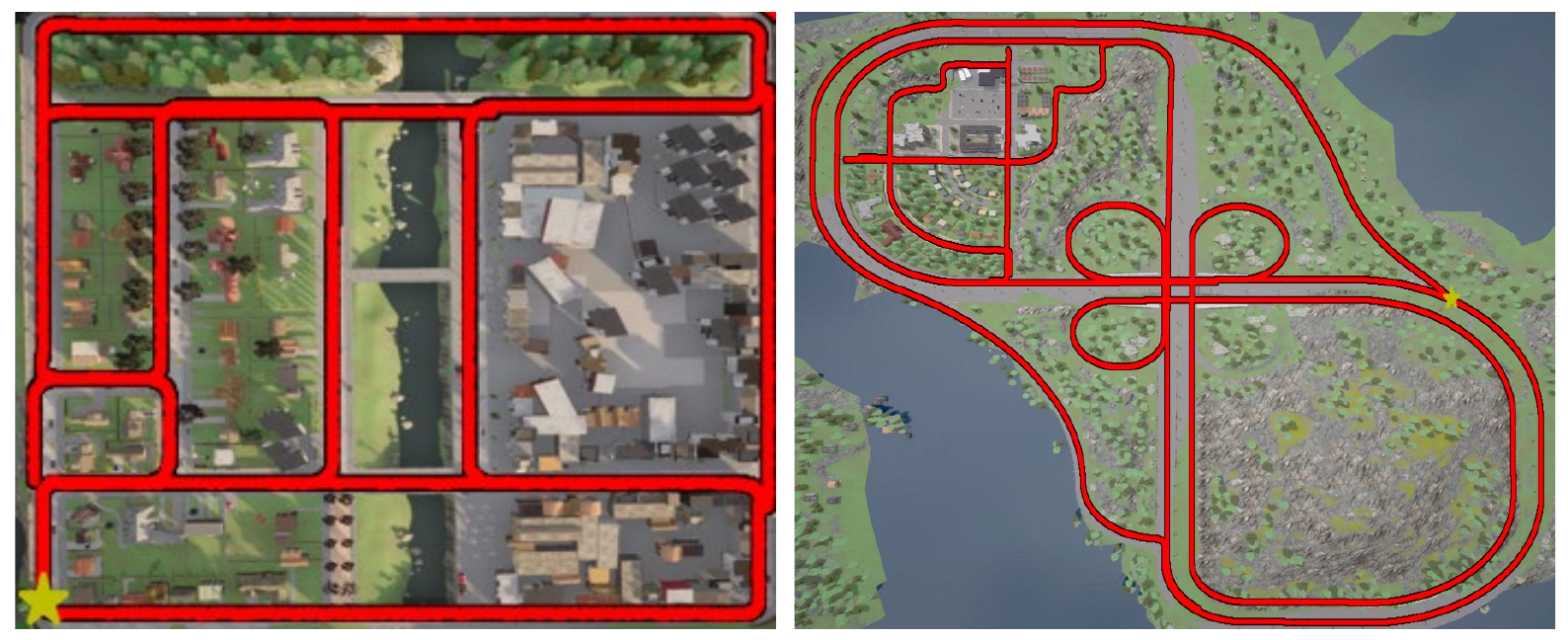} % Reduce the figure size so that it is slightly narrower than the column. Don't use precise values for figure width. This setup will avoid overfull boxes.
\caption{Trajectory map of Town1Loc and Town4Loc. The yellow star indicates the starting point of the traversal.}
\label{tra}
\end{figure}

\begin{figure*}[t]
\centering
\includegraphics[width=\textwidth]{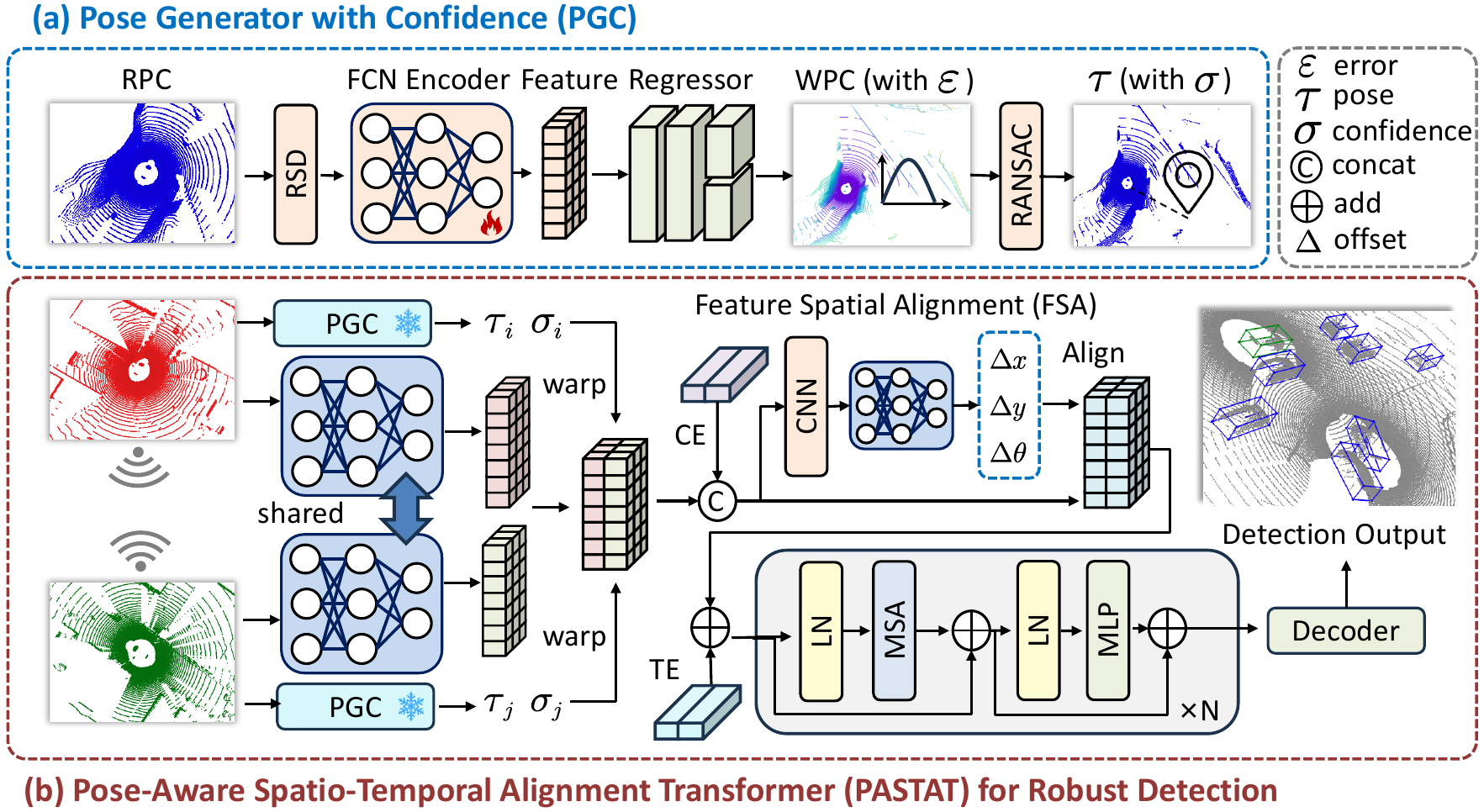} % Reduce the figure size so that it is slightly narrower than the column.
\caption{The overall architecture of the proposed framework. The architecture consists of two modules: (a) Pose Generator with Confidence (PGC), (b) Pose-Aware Spatio-Temporal Alignment Transformer (PASTAT). The agent obtains the global pose and confidence through PGC, and then aligns the features through PASTAT to obtain the collaborative detection result. \textit{RSD}: Redundant Sample Downsampling; \textit{RPC}: Raw LiDAR Point Cloud; \textit{WPC}: Point Cloud in World coordinate system; \textit{CE}: Confidence Embedding; \textit{TE}: Temporal Encoding.}
\label{method}
\end{figure*}

The data collection for V2VLoc is performed using OpenCDA \cite{opencda}, a collaborative simulation platform based on CARLA \cite{carla}. This platform facilitates the simulation of collaborative driving scenarios involving multiple agents. 
Town1Loc and Town4Loc are constructed from the Town1 and Town4 maps of CARLA using a single agent for localization tasks. We visualize the agent's trajectory, as shown in Fig. \ref{tra}, where Town1Loc and Town4Loc cover 31.31 km of roads that include underbridges, interchanges, roundabouts, ramps, gas stations, urban streets, and so on. V2VDet provides a variety of collaborative driving scenarios on Town1 and Town4 maps involving 2 to 4 agents per scene. The subset comprises 11,598 LiDAR and 4D-radar frames, 46,392 camera images, and 260,210 annotated vehicle 3D  bounding boxes. 
% \end{itemize}

\subsection{Sensor Setup}
Each vehicle in the V2VLoc dataset is equipped with multiple sensors to ensure comprehensive perception. The sensor suite includes:
\begin{itemize}
    \item Cameras: Four RGB cameras are mounted on each vehicle, positioned at (2.5, 0.1, 0.0), (0.0, 0.3, 1.8, 100), (0.0, –0.3, 1.8, –100), and (–2.0, 0.0, 1.5, 180), providing full 360° visual coverage around the vehicle.
    \item LiDAR: A single 64-channel LiDAR sensor is used, with a maximum range of 120 meters and a vertical field of view from –25° to 2°. It operates at a rotation frequency of 20 Hz and has a noise standard deviation of 0.02.
    \item 4D Radar: The radar has a sensing range of 150 meters, a 120° horizontal FOV, and a 30° vertical FOV, offering complementary motion and depth information under various weather and lighting conditions.
    \item GPS and IMU: The vehicle GNSS has an altitude noise of 0.001 meters. The vehicle IMU includes heading noise of 0.1°, and speed noise of 0.2 m/s. The RSU (Road Side Unit) GNSS provides altitude measurements with a noise level of 0.05 meters.
\end{itemize}

% \subsection{Trajectory map of Town1Loc and Town4Loc}

\section{Method}

\subsection{Problem formulation}
In GNSS-denied environments, agents are unable to acquire accurate pose information via external localization systems. Therefore, unlike traditional collaborative detection frameworks, for each agent $i$, our collaborative detection process proceeds as follows:
\begin{equation}
\tau _i, \sigma _i=\mathcal{P} \left( O_i \right) 
\end{equation}
\begin{equation}
F_{i}^{*}=\varPsi  \cdot \mathcal{S}(\sigma) \cdot \mathcal{A} \left( f\left( O_i \right) ,f\left(\{O_j\}_{j \in \mathcal{N}}\right), \tau \right)
\end{equation}
\begin{equation}
\hat{B}_i=D\left( F_{i}^{*} \right) 
\end{equation}

In Eq. (1), each agent first estimates its own pose $\tau_i$ and confidence $\sigma_i$ from the local observation $O_i$ using the pose estimation module $\mathcal{P}$. In Eq. (2), the feature extraction function $f(\cdot)$ encodes the local point cloud $O_i$ into features $f(O_i)$. These features are then coarsely aligned with the encoded features of neighboring agents $f\left(\{O_j\}_{j \in \mathcal{N}}\right)$ using the coarse alignment module $\mathcal{A}$ by $\tau$. Subsequently, a confidence-aware spatial alignment module $\mathcal{S}$ and a temporal transformer encoder $\varPsi$ are applied to yield the aggregated feature $F_i^*$. Finally, In Eq. (3), the detection head $D$ takes $F_i^*$ as input to produce the final detection results $\hat{B}_i$.

\subsection{Pose Generator with Confidence (PGC)}

To effectively overcome the limitation of the graph matching methods \cite{icp, freealign} that require agents to share multiple commonly observed objects, a naive idea is to allow each agent to directly learn the correspondence between its observations and its pose by capturing the underlying geometric and semantic structure of the scene. 
% This facilitates more accurate pose estimation and spatial feature alignment.

Inspired by this idea, we adopt the architecture of regression-based LiDAR localization, which estimates the 6-DoF pose by learning to transform the raw LiDAR point cloud (RPC) into the world-frame point cloud (WPC) and matching points between them using RANSAC \cite{ransac}. However, as shown in the left figure of Fig. \ref{fig2}, regression-based LiDAR localization still suffers from pose estimation errors. Therefore, we propose the PGC module, which jointly estimates the pose and its confidence to enable more reliable collaborative alignment.
% we adopt the LightLoc \cite{lightloc} architecture, which learn to transform the Raw LiDAR Point Cloud (RPC) into the World-frame Point Cloud (WPC). By this way, LightLoc estimates the 6-DoF pose by matching points in the RPC ($p_r$) with those in the WPC ($p_w$). However, LightLoc \cite{lightloc} still suffers from pose estimation errors in challenging scenarios. Therefore, we propose the PGC module, which jointly estimates the pose and its confidence to enable more reliable collaborative alignment. for a given input point cloud $P\in \mathbb{R} ^{N\times 3}$,

Specifically, we learn a mapping from the input raw point cloud to global scene coordinates $y_i$, the network regresses $y_i$ by minimizing the average Euclidean distance to the ground truth $y_i^*$, as shown in Eq. (4).
% Specifically, we define the pose as the transformation that maps the RPC to the WPC:
% \begin{equation}
% p_w=T_{r\rightarrow w}\cdot p_r
% \end{equation}
% where $T$ denotes a 4×4 pose transformation matrix. Then, by learning a mapping from the input point cloud to global scene coordinates $y_i$, the network regresses $y_i$ by minimizing the average Euclidean distance to the ground truth $y_i^*$, as shown in (5).
% by learning $\mathcal{Y} =f\left( P \right)$, a mapping from the input point cloud, the network regresses the corresponding scene coordinates under the global pose, denoted as $y_i\in \mathcal{Y}$. This is achieved by minimizing the average Euclidean distance between the predicted scene coordinates $y_i$ and the ground truth $y_{i}^{*}$ as shown in (5).
\begin{equation}
u_i=||y_i-y_{i}^{*}||_1
\end{equation}

Meanwhile, we further propose to let the network learn to predict the pose error $\varepsilon$. The overall loss function is defined as follows:

\begin{equation}
\mathcal{L} _{reg}=\frac{1}{|\mathcal{Y} |}\sum_{y_i\in \mathcal{Y}}{u_i+||u_i-\varepsilon _i||_1}
\end{equation}

where $\mathcal{Y}$ is the set of samples, and $|\mathcal{Y}|$ is the number of samples.
And the pose confidence is derived from Eq. (6):
\begin{equation}
\sigma _i=\frac{1}{1+\varepsilon _{i}^{2}}
\end{equation}

A lower pose error results in a higher confidence, and a higher pose error results in a lower confidence, ensuring that more reliable pose estimates contribute more in downstream tasks such as feature alignment or fusion. Then, RANSAC \cite{ransac} is used to obtain the final poses and confidences.

Moreover, to accelerate network training, we adopt the Redundant Sample Downsampling (RSD) strategy \cite{lightloc}, which effectively reduces the number of redundant points on the input. This not only reduces the model parameter count but also enhances the training efficiency by focusing computation on informative samples.

\subsection{Pose-Aware Spatio-Temporal Alignment Transformer (PASTAT)}
% As illustrated in Fig. \ref{method}, each agent obtains its estimated pose and confidence via a frozen PGC module, which allows coarse feature alignment of visual features extracted by the feature encoder. However, due to inherent global pose errors introduced by PGC, the features cannot be perfectly aligned at this stage. To address this, we introduce a learnable Pose-Aware Spatio-Temporal Alignment Transformer (PASTAT).
PASTAT consists mainly of the following components: (i) Feature Extraction and Coarse Alignment; (ii) Feature Spatial Alignment (FSA); (iii) Vision Transformer with Temporal Encoding; (iv) Decoder and Loss.
\subsubsection{Feature Extraction and Coarse Alignment.} 
Each agent $i$ receives a raw observation of point cloud $O_i$ as input. We use a shared feature encoder to extract semantically meaningful spatial features.
% The encoder can be instantiated as a BEV-based CNN or a sparse point cloud transformer to capture geometric and semantic structures from raw LiDAR data effectively.
% Then we define a communication graph $\mathcal{G} = \left(\mathcal{V}, \mathcal{E}\right)$, where the nodes $\mathcal{V}$ represent the agents, and the edges $\mathcal{E}$ denote the communication links. In this graph, the ego agent acts as the requester, while other vehicles within its communication range act as supporters. 
With PGC, each agent $j \in \mathcal{N}$ transmits its encoded features $F_j$, the estimated pose $\tau_j \in \text{SE}(3)$, and the corresponding confidence score $\sigma_j \in [0, 1]$ to the ego agent. With estimated poses $\tau$, we first compute the relative transformation from neighbor $j$ to ego agent $i$ and apply a rigid transformation to each neighbor's feature map to concatenate all features. Note that the features here are not fully aligned due to posture errors and have not been fused.

\subsubsection{Feature Spatial Alignment (FSA).}
Although the coarsely aligned features provide a basic foundation for multi-agent fusion, they may still suffer from global pose noise, leading to sub-optimal detection performance. To address this limitation, we introduce the FSA module to enable precise feature-level alignment.

The core idea of FSA is to design a learnable alignment network that estimates inter-agent feature misalignments. Specifically, we first apply a normalization operation to the confidence scores $\sigma$ generated by the PGC module, referred to as Confidence Embedding (CE), and then concatenate the normalized confidence with the extracted feature maps, as shown in Eq. (7). This enables the network to explicitly incorporate pose reliability into the alignment process. 
\begin{equation}
F_{i}^{\prime}=F_i\oplus \frac{\sigma _i}{\sum_j^N{\sigma _j}}
\end{equation}

To achieve accurate alignment, we introduce a dedicated alignment network to predict spatial transformation parameters $\Delta T_{j\rightarrow i}\in \mathbb{R} ^K$, which is defined with $K$ degrees of freedom. In our design, we adopt a 3-DoF transformation, where the predicted offset includes translation in the $x$ and $y$ directions, and a rotation angle $\theta$. Our network uses convolutional neural networks (CNNs) and fully connected layers to learn the offsets of the feature shift between agents. These offsets are then used to adjust the spatial positions of the fused features accordingly.

% To supervise the training of the alignment network, we do not use ground-truth pose information to supervise the learning of the alignment network. Instead, we use the ground-truth bounding boxes as a supervisory signal. By minimizing the discrepancy between the aligned features and the true object locations, the network learns to compensate for pose-induced misalignments in a data-driven manner. We argue that accurate feature matching leads directly to higher-quality detection performance, especially under challenging pose uncertainty.

\subsubsection{Vision Transformer with Temporal Encoding.}
We adopt a Vision Transformer (ViT) encoder \cite{v2x-vit} for global context modeling, treating the aligned feature map as token sequences. Each transformer layer comprises Multi-Head Self-Attention (MSA) and a Feed-Forward Network (MLP), both with Layer Normalization (LN) and residual connections. Given input $z^{(l-1)}$ to the $l^{th}$ layer, the computations are defined as:
\begin{equation}
z^{\left( l \right) ^{\prime}}=\text{MSA}\left( \text{LN}\left( z^{\left( l-1 \right)} \right) \right) +z^{\left( l-1 \right)}
\end{equation}
\begin{equation}
z^{\left( l \right)}=\text{MLP}\left( \text{LN}\left( z^{\left( l \right) ^{\prime}} \right) \right) +z^{\left( l \right) ^{\prime}}
\end{equation}

To enable temporal awareness, we also incorporate a temporal encoding scheme \cite{v2x-vit} into the input of the transformer. Concretely, given a sequence of aligned features from multiple time steps $\{ \tilde{F}_i^{(t)} \}_{t=1}^T$, we inject a learnable or sinusoidal temporal encoding $E_t$ into each feature token before feeding it into the Transformer:
\begin{equation}
z^{(0)}_t = \text{Flatten}(\tilde{F}_i^{(t)}) + E_t 
\end{equation}
where $E_t \in \mathbb{R}^{D}$ represents the temporal encoding corresponding to the time step $t \in \{1, 2, \ldots, T\}$, and $D$ is the dimension of the embedding of the token. For each time step $t$ , the temporal encoding vector $E_t$ is defined as:
\begin{equation}
E_t^{(2k)} = \sin\left(\frac{t}{10000^{(2k)/D}}\right) \quad 
\end{equation}
\begin{equation}
E_t^{(2k+1)} = \cos\left(\frac{t}{10000^{(2k+1)/D}}\right)
\end{equation}
where $k = 0, 1, \ldots, D/2 - 1$.

% By assigning unique encoding vectors to each frame or time step, the model can perceive and distinguish data from different moments, thereby effectively modeling inter-frame dynamics and temporal dependencies.

\begin{table*}[!t]
\centering
\renewcommand{\arraystretch}{1.1}
% \resizebox{0.8\textwidth}{!}{
\begin{tabular}{c|c|ccc|ccc}
\hline
\multirow{2}{*}{Method} & \multirow{2}{*}{Reference} & \multicolumn{3}{c|}{V2V4Real} & \multicolumn{3}{c}{V2VDet} \\
                        &                            & AP@0.3   & AP@0.5   & AP@0.7  & AP@0.3  & AP@0.5  & AP@0.7 \\ \hline
No Fusion                  & -                          & 47.50    & 39.83    & 22.02   & 40.16   & 37.28   & 22.31  \\
Late Fusion                    & -                          & 40.18    & 34.60    & 15.64   & 50.88   & 47.53   & 29.58  \\
Where2comm \cite{where2comm}             & NeurIPS2022                & 61.30    & 57.61    & 37.75   & 57.80   & 49.38   & 33.95  \\
CoBEVT \cite{cobevt}                 & CORL2022                   & 59.03    & 56.11    & 34.69   & 63.68   & 59.50   & 39.11  \\
V2X-ViT \cite{v2x-vit}                & ECCV2022                   & 60.15    & 56.90    & 35.84   & 67.52   & 62.70   & 43.46  \\
CoAlign \cite{CoAlign}                & ICRA2023                   & 62.11    & 58.93    & 34.38   & 66.75   & 62.98   & 37.50  \\
ERMVP   \cite{ermvp}                & CVPR2024                   & 60.86    & 58.90    & 38.74   & 67.86   & 64.24   & 47.13  \\
TraF-Align \cite{traf-align}             & CVPR2025                   & 62.11    & 56.11    & 31.54   & 72.65   & 66.75   & 48.29  \\
PASTAT (Ours)            & -                          & \textbf{63.52}    & \textbf{61.51}    & \textbf{40.29}  & \textbf{76.97}   & \textbf{71.15}   & \textbf{52.55}  \\ \hline
\end{tabular}
% }
\caption{Performance comparison of vehicle class on V2V4Real \cite{v2v4real} test and V2VDet test dataset. The results are reported in AP@0.3 / 0.5 / 0.7, and the agent uses the ground truth pose with 1.0 / 1.0 noise level (m/°) in V2V4Real \cite{v2v4real}, while PGC is used to obtain the pose in V2VDet.}
\label{table1}
\end{table*}

\subsubsection{Decoder and Loss.}
Based on the final fused feature representation, we generate the detection outputs using a detection decoder. Each predicted bounding box $\hat{b}_i$ represents a rotated 3D object, parameterized as:
$\hat{b}_i = (x, y, z, h, w, l, \theta) $.
Following prior work \cite{PointPillars}, we adopt the Smooth $L1$ loss for bounding box regression and the Focal Loss \cite{FocalLoss} for classification, which together ensure robust training under class imbalance and spatial uncertainty.

% where  $(x, y, z)$ denotes the 3D position of the object center, $(h, w, l)$ represents the object dimensions (height, width, length), and $\theta$ is the rotation angle around the vertical axis.
% These predicted objects constitute the final output of the proposed collaborative perception system.
\begin{table}[t]
\centering
\renewcommand{\arraystretch}{1.1}
\setlength{\tabcolsep}{1mm}
% \resizebox{1\columnwidth}{!}{
\begin{tabular}{c|ccc}
\hline
Method    & Time (s) & $\mathcal{C}$ (log2)   & $\delta_s$ (\%)\\ \hline
ICP \cite{icp}      & 0.6009    & \multirow{2}{*}{7.37}     &    0.813        \\
FreeAlign \cite{freealign}  & 0.0878  &                          & 60.26            \\ \hline
HypLiLoc \cite{wang2023hypliloc} & 0.5652  &  \multirow{4}{*}{4.62}                        & 80.96            \\ 
SGLoc  \cite{sgloc}    & 0.0242  &   &       83.47     \\
DiffLoc   \cite{diffloc}    & 0.1911  &   &       81.32 \\
LightLoc \cite{lightloc}  & 0.0081  &           & 98.35  \\ 
 \hline
\end{tabular}
% }
\caption{Comparison of Alignment Performance. We compared alignment results in the V2VDet dataset using ground truth bounding boxes for ICP and FreeAlign. Time represents the average alignment time, $\mathcal{C}$ represents the communication volume during alignment, $\delta_s$ (\%) indicates the success rate, which means translation error is less than 3m.}
\label{intro1}
\end{table}

\section{Experiments}

\subsection{Dataset and Evaluation Metrics}

% comprising three subsets: Town1Loc, Town4Loc, and V2VDet.
% (i) Town1Loc and Town4Loc are built on the CARLA simulator’s Town1 and Town4 maps, covering 31.31 km of roads that include tunnels, interchanges, roundabouts, ramps, gas stations, and urban streets. We further introduce diverse weather simulations (sunny, snow\cite{snow}, fog\cite{fog}) to increase environmental diversity and validate the robustness of SCR-based localization. (ii) V2VDet includes a variety of collaborative scenarios with LiDAR, camera, and 4D radar inputs, totaling 11,598 radar frames, 46,392 images, and 260,210 vehicle bounding boxes.

We conduct experiments on both our V2VLoc dataset and the real-world V2V4Real \cite{v2v4real} dataset. 
For V2VLoc, we use 6,697 training, 2,017 validation, and 2,884 test frames for collaborative 3D object detection.
For V2V4Real \cite{v2v4real}, it is the first large-scale real-world dataset for vehicle-to-vehicle (V2V) collaborative perception. It covers 410 kilometers of driving, providing over 20K frames of LiDAR data, with 14,210 frames for training, 2,000 for validation, and 3,986 for testing. To ensure fair and consistent evaluation, we adopt the official metrics used in previous works, including mean Average Precision (mAP) under IoU thresholds of 0.3, 0.5, and 0.7.

\subsection{Implementation Details}
For the PGC module, we implement our method using PyTorch and train the model on a multi-GPU server with 24 data loading workers. 
% We use four Town1Loc and three Town4Loc trajectories for training and all V2VDet trajectories for testing. 
The model is optimized using the AdamW optimizer with an initial learning rate of 1e-3 and a weight decay of 1200. We adopt a total of 100 training epochs with a batch size of 100. 

For the PASTAT module, we train the network on the V2VDet dataset using the OpenCOOD \cite{opv2v} framework and adopt PointPillar \cite{PointPillars} as the lightweight backbone model. 
Models are trained for 60 epochs with a batch size of 2 using the Adam optimizer (initial LR = 0.001, weight decay=1e-4) and a MultiStep scheduler (decay at epochs 15 and 50). All experiments were trained on 4 NVIDIA GeForce RTX 3090 GPUs.

\begin{table*}[t]
\centering
\setlength{\tabcolsep}{1mm}
\renewcommand{\arraystretch}{1.2}
% \resizebox{1.0\textwidth}{!}{
\begin{tabular}{c|c|c|c|c|c}
\hline
Method/Metric          & \multicolumn{5}{c}{AP@0.3/0.5/0.7}                                                                \\ \hline 
Noise   Level $\sigma_t/\sigma_r$ (m/°) & 0.0/0.0           & 1.0/1.0           & 2.0/2.0           & 3.0/3.0           & 4.0/4.0           \\ \hline
No Fusion               & \multicolumn{5}{c}{40.16/37.28/22.31}                                                             \\ \hline
Early Fusion                                  & 62.95/61.05/48.43 & 50.02/44.59/27.05 & 48.81/46.55/32.65 & 50.60/48.50/34.60  & 51.33/49.21/35.25 \\
Late Fusion                                  & 58.03/57.03/47.15 & 49.29/39.33/27.27 & 50.58/49.57/40.78 & 50.68/49.54/40.77 & 50.60/49.53/40.68 \\
CoBEVT                                       & 65.12/62.59/43.86 & 61.21/57.22/37.71 & 58.30/55.54/37.94 & 57.95/56.03/39.17 & 59.24/57.48/40.91 \\
Where2comm            & 68.77/59.89/36.64 & 66.11/57.78/35.35 & 63.29/55.83/34.45 & 63.77/57.59/36.49 & 65.36/59.35/38.05 \\
Coalign              & 68.52/65.41/43.24 & 58.14/52.48/31.07 & 57.07/53.66/31.96 & 57.27/53.47/31.00 & 57.18/53.77/32.05 \\
V2X-ViT             & 73.14/69.11/48.69 & 63.59/56.61/27.56 & 61.13/56.74/27.78 & 61.30/57.76/29.71 & 61.29/57.93/30.14 \\
ERMVP                 & 69.95/69.00/57.93 & 55.50/50.64/27.28 & 45.77/40.73/30.32 & 42.41/41.60/32.88 & 42.69/42.10/33.84 \\
Traf-Align         & 78.24/73.08/42.38 & 72.19/59.51/36.26 & 65.82/56.08/31.60  & 62.98/55.23/32.30 & 62.49/56.86/33.90 \\ \hline
PASTAT(Ours)                   & \multicolumn{5}{c}{\textbf{76.97/71.15/52.55}}                                                             \\ \hline
% PASTAT(Ours)                   & \textbf{76.97/71.15/52.55} & \textbf{76.97/71.15/52.55} & \textbf{76.97/71.15/52.55}  & \textbf{76.97/71.15/52.55} & \textbf{76.97/71.15/52.55}                                                       \\ \hline
\end{tabular}

\caption{Performance comparison of different models on the V2VDet dataset under different GNSS perturbations. We evaluate AP@0.3/0.5/0.7 of different fusion methods, and our modules consistently maintain strong performance as they are not affected by GNSS disturbances.}
\label{table3}
\end{table*}

\subsection{Quantitative Evaluation}

% \subsubsection{Comparison of Localization Performance.}
% As shown in Tab. \ref{intro1}, regression-based localization methods outperform traditional and graph-matching-based approaches. LightLoc \cite{lightloc} achieves the highest alignment success rate, 98.35\%, with low communication cost and fastest runtime. 

% provides a similarly strong trade-off. These results demonstrate that regression-based methods offer a superior balance of accuracy, efficiency, and communication overhead.
\subsubsection{Comparison of Alignment Performance.}
As shown in Tab. \ref{intro1}, regression-based localization methods clearly outperform traditional approaches. LightLoc \cite{lightloc} achieves the highest alignment success rate, 98.35\%, with a low communication volume of 4.62 and a fast runtime of 0.0081s, which demonstrates the effectiveness of LiDAR localization methods for collaborative feature alignment in GNSS-denied environments. 

% Our PGC is mostly based on LightLoc \cite{lightloc} and achieved the same performance on the V2VDet dataset.
\subsubsection{Comparison of Detection Performance.}
Tab. \ref{table1} presents the comparison of the 3D object detection performance of vehicle class between models in the V2V4Real \cite{v2v4real} and V2VDet datasets. 
% We achieved the best performance based on LightLoc\cite{lightloc} with an unfrozen backbone.
% As shown in Tab. \ref{table1}, "Single" refers to single-agent detection without any collaboration, while "Late" indicates methods where agents only share detected bounding boxes and utilize non-maximum suppression to produce the final results. The intermediate fusion methods include Where2comm\cite{where2comm}, CoBEVT\cite{cobevt}, V2X-ViT\cite{v2x-vit}, CoAlign\cite{CoAlign}, ERMVP\cite{ermvp}, and TraF-Align\cite{traf-align}, covering current state-of-the-art approaches. 

For the V2VDet dataset, we simulate a GNSS-denied environment, where all fusion methods utilize the PGC module to generate poses. Our proposed PASTAT achieves the best performance, outperforming the previous state-of-the-art by 4.40\% and 4.26\% on AP@0.5 and AP@0.7, respectively. 

For the V2V4Real ~\cite{v2v4real} dataset, we do not train our PGC module on it, as it contains only a single LiDAR traversal per location. This setup violates a key assumption of regression-based LiDAR localization: training and testing must occur on distinct traversals to prevent overfitting to scene-specific geometry. Therefore, to fairly benchmark downstream collaborative detection performance, ground truth poses with 1.0 / 1.0 noise level (m/°) \cite{where2comm, ermvp, freealign} are used, with pose confidence obtained from pose error. It can be seen that the PASTAT module can effectively solve the problem of inaccurate LiDAR pose in both real and simulated datasets through feature alignment.
% \begin{figure}[t]
% \centering
% \includegraphics[width=1.0\columnwidth]{figure/err} % Reduce the figure size so that it is slightly narrower than the column. Don't use precise values for figure width. This setup will avoid overfull boxes.
% \caption{Robustness performance curves on the V2VDet dataset at AP@0.5 and AP@0.7.
% % During the test phase, as the level of Gaussian perturbation increases, other baseline models exhibit a clear performance drop, while our PGC-based PASTAT model maintains consistently high performance.
% }
% \label{fig4}
% \end{figure}

\subsubsection{Robustness to Pose Errors by GNSS.}
To assess robustness under GNSS-denied conditions, we simulate varying GNSS noise levels and evaluate their impact on different collaborative detection pipelines. Specifically, we simulate varying levels of localization noise by adjusting the standard deviation of Gaussian pose perturbations, ranging from 0.0/0.0 to 4.0/4.0 (m/°). The performance of different models under these conditions is evaluated in terms of AP@0.3/0.5/0.7 (see Tab. \ref{table3}). While all models exhibit a noticeable decline in performance as the noise level increases, the degree of robustness varies significantly among them. Notably, our PASTAT model, which incorporates pose generation via the PGC module, demonstrates superior robustness to GNSS localization perturbations.
% 具体而言，如表2所示，我们通过改变Pose Error std来实现不同的Noise level的设置，测试从0.0/0.0到4.0/4.0的位姿高斯扰动下，不同的模型对应的AP0.3/0.5/0.7。并绘制了在AP@0.5和AP@0.7下的不同模型随着Noise Level变化的曲线图，如图4可以看出，不同模型对于噪声的鲁棒性各不相同，但均呈现明显的下降趋势，而我们的基于PGC位姿生成的PASTAT模型呈现出较好的鲁棒性。

\begin{table}[t]
\centering
% \resizebox{1\columnwidth}{!}{
\setlength{\tabcolsep}{1mm}
\renewcommand{\arraystretch}{1.1}
\begin{tabular}{c|c|ccc|ccc}
\hline
\multirow{2}{*}{ID} & \multirow{2}{*}{PGC} & \multicolumn{3}{c|}{PASTAT} & \multicolumn{3}{c}{V2VLoc} \\
                    &                      & CE     & FSA     & TE     & AP@0.3  & AP@0.5  & AP@0.7 \\ \hline
1                   &                      &        &           &        & 42.53   & 38.49   & 24.27  \\
2                   & \checkmark                    &        &           &        & 57.95   & 52.68   & 39.69  \\
3                   & \checkmark                     & \checkmark       &           &        & 69.43   & 63.43   & 41.07  \\
4                   & \checkmark                     & \checkmark       & \checkmark          &        & 72.82   & 66.70   & 48.19  \\
5                   & \checkmark                     & \checkmark       & \checkmark          & \checkmark       & \textbf{76.97}   & \textbf{71.15}   & \textbf{52.55}  \\ \hline
\end{tabular}
% }
\caption{Results of the ablation study of the core components on the V2VLoc dataset. We use PointPillars \cite{PointPillars} and ViT \cite{v2x-vit} as baselines and gradually add our modules to verify the performance of our method.}
\label{table2}
\end{table}

\subsection{Ablation Studies}
To evaluate the individual contributions of each component in our proposed framework, we conduct a comprehensive ablation study on the V2VLoc dataset. The ablation focuses on four key modules: Pose Generator with Confidence (PGC), Confidence Embedding (CE), Feature Spatial Alignment (FSA), and Temporal Encoding (TE). The corresponding experimental results are summarized in Tab. \ref{table2}. We take the PointPillars \cite{PointPillars} and Vision Transformer Encoding \cite{v2x-vit} as our baseline and progressively add the following modules to evaluate the contribution of each component: (i) PGC, (ii) CE, (iii) FSA, and (iv) TE. The FSA does not include the CE module in this context. As shown in Tab. \ref{table2}, all modules contribute positively to performance improvement. In particular, the proposed PGC module improves AP@0.3/0.5/0.7 by 15.42\%/14.19\%/15.42\%, respectively, while the PASTAT module further boosts AP@0.3/0.5/0.7 by 19.02\%/18.47\%/12.86\%, respectively.

\section{Conclusion}
In this paper, we propose a novel GNSS-free collaborative perception framework that enables robust multi-agent detection without GNSS signals. By leveraging LiDAR localization, our method establishes accurate 3D correspondences between local observations and a shared global scene, allowing each agent to estimate its pose without relying on external localization systems. We propose a new dataset, V2VLoc, and to enhance the robustness and accuracy of the fusion process, we further incorporate modules such as PGC and PASTAT. Extensive experiments on the V2VLoc and V2V4Real demonstrate the effectiveness of our approach. 
% \subsubsection{Limitations and Future Work.}While our work lays a solid foundation, collaborative perception under GNSS-denied conditions remains an underexplored area. Effectively leveraging scene information for pose regression to enable accurate multi-agent fusion continues to be a promising and important direction for future research.

\section{Acknowledgments}
This work was supported in part
 by the National Natural Science Foundation of China
 (No.42571514).
\bibliography{main}

\end{document}